\documentclass[letterpaper,10pt,twocolumn]{article}

\usepackage[top=0.75in,bottom=0.75in,left=0.625in,right=0.625in]{geometry}
\usepackage{amsmath}
\usepackage{amsfonts}
\usepackage{siunitx}
\usepackage{graphicx}
\usepackage{multirow}
\usepackage{hhline}
\usepackage{adjustbox}
\usepackage{booktabs}
\usepackage{algorithm}
\usepackage{algpseudocode}
\usepackage{subcaption}
\usepackage[numbers,sort&compress]{natbib}
\usepackage{doi}
\usepackage{xcolor}
\usepackage{xspace}
\usepackage{tikz}
\usetikzlibrary{positioning,arrows.meta,calc,backgrounds,fit,shapes.geometric}
\usepackage{hyperref}
\usepackage[capitalise]{cleveref}
\usepackage{nameref}

\hypersetup{
  unicode=true,
  bookmarks=true,
  bookmarksnumbered=true,
  colorlinks=true,
  linkcolor=blue,
  citecolor=blue,
  urlcolor=blue,
  pdftitle={Body-Grounded Replanning for Physically Adaptive Manipulation},
  pdfauthor={Namiko Saito and Hiroshi Kera}
}

\newcommand{\pms}[1]{{\scriptstyle\,\pm#1}}
\newcommand{\paraDraft}[1]{}

\newcommand{\remove}[1]{}
\newcommand{\namiko}[1]{}
\newcommand{\revise}[1]{#1}

\title{Body-Grounded Replanning for Physically Adaptive Manipulation}

\author{
  Namiko Saito$^{1,2}$ \qquad Hiroshi Kera$^{3,4}$\thanks{Corresponding author.}\\[0.5em]
  \normalsize $^1$Microsoft Research Asia - Tokyo \quad $^2$Waseda University\\
  \normalsize $^3$Chiba University \quad $^4$National Institute of Informatics\\
  \normalsize\texttt{namikosaito@microsoft.com \quad kera@chiba-u.jp}
}

\date{}

\begin{document}

\maketitle

\begin{abstract}
Manipulation requires not only reasoning about the external environment, but also about the robot's physical condition.
A strategy may remain geometrically feasible while becoming physically unsuitable due to increased joint load or limited mobility, yet internal physical state is typically used only for low-level control.
We propose \emph{body-grounded high-level replanning}, which uses internal physical state to~\namiko{deleted: reconsider and} adapt manipulation strategies during execution.
Body-state events trigger strategy replanning, and an LLM interprets the underlying joint-level state, recent execution statistics, and execution history to select a context-dependent alternative, while leaving the task objective and low-level controller unchanged.
We evaluate the framework on a reaching task under controlled load and asymmetric mobility constraints in simulation and on a real robot.
Our experiments show that body-grounded replanning maintains high task success while reducing physical effort and enabling more efficient strategy adaptation.
Additional contact-rich manipulation experiments demonstrate the applicability of the same replanning interface beyond reaching.
These results show that internal physical state can inform not only low-level control, but also high-level decisions about how a manipulation task should be performed.
The video is available at \url{https://youtu.be/_qclDPyHm4U}.
\end{abstract}

\section{Introduction}
\label{sec:introduction}
\begin{figure}[t]
\centering
\includegraphics[width=\linewidth]{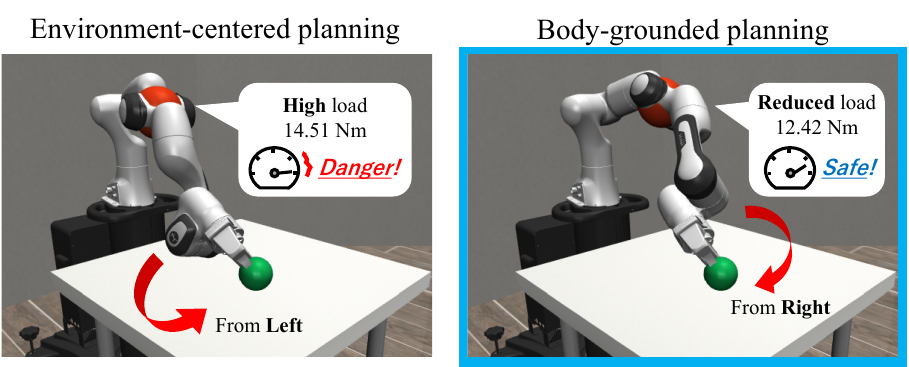}
\caption{\namiko{changed fig}
Environment-centered and body-grounded strategies reach the same task-space goal, but the body-grounded strategy reduces load on the affected joint (red, Joint 4) by selecting an alternative approach.
}
\vspace{-2mm}
\label{fig:concept}
\end{figure}

Manipulation requires reasoning not only about the external world but also about the robot's own physical condition.
\revise{Importantly, the physical condition sometimes becomes evident only after execution begins; an arm might not respond properly due to a malfunctioning joint or its motion might be unexpectedly impeded by an obstacle encountered during execution.
In such a case, humans would adaptively change their \textit{strategies}, e.g., finding another direction of approach that avoids pain or an obstacle, or switching to another arm. 
Determining a favorable strategy requires adaptive, sequential trials of strategies, with a high-level analysis of the local successes and failures.
Figure~\ref{fig:concept} shows a reaching task approached using two different strategies. 
Both are geometrically feasible, but one imposes a high load on a joint, while in the other, the robot senses the increased load and changes the arm configuration to reduce it. 
}

\revise{However, existing high-level planning primarily uses scene observations, object geometry, and task specifications to determine what to do and how to approach a task before execution~\cite{ichter2022saycan, codeaspolicies2022}, and also during execution~\cite{Bhat_2025, mei2024replanvlm, AgiaSinhaEtAl2024, liu2023reflect}.
Internal physical state is mainly used for low-level control~\cite{khatib1987operational, siciliano2010robotics} isolated from higher-level planning.
}
\revise{
Consequently, a robot may persist with a geometrically valid strategy even after its own physical feedback indicates that another strategy would be more appropriate.
This motivates closing the loop between physical experience during execution and high-level strategy selection.
}

In this study, we address this gap through \emph{body-grounded high-level replanning}, which closes a strategy-level feedback loop from the robot's internal physical state to its manipulation strategy.
Body-state events indicate when the current strategy \revise{has become physically unfavorable}.
Once triggered, an LLM interprets the underlying joint-level state together with recent execution statistics and execution history to select a physically suitable alternative, while leaving the low-level controller unchanged.

A key motivation for this design is that detecting physical difficulty does not by itself determine how the strategy should change.
For example, high joint effort may indicate that the current strategy is undesirable, but it does not specify whether the robot should approach from the left, right, or above.
Likewise, proximity to a joint limit may require different responses depending on which joint is constrained, the current configuration, and the strategies already attempted.
Rather than prescribing a fixed response to each body-state event, our approach uses the underlying physical state and execution context to select among available strategies.

We evaluate body-grounded replanning on a reaching task in which the external task and environment remain fixed while the robot's physical condition changes through increased load and asymmetric mobility constraints.
Experiments in simulation and on a real robot compare our approach with fixed planning, body-agnostic replanning, and rule-based body replanning.
\revise{Across controlled load and mobility variations, body-grounded replanning maintains high task success rate while reducing physical execution cost and enabling more efficient strategy adaptation.}
We further apply the same body-grounded replanning interface to three contact-rich manipulation tasks beyond reaching.

The contributions of this work are:
\begin{itemize}
    \item We formulate \emph{body-grounded high-level replanning}, in which internal physical state \revise{during execution} is used to evaluate and adapt high-level manipulation strategies even when the task remains geometrically feasible.

    \item We introduce a strategy-level replanning framework that uses body-state events to \revise{determine when a strategy change is needed, while contextual reasoning determines} how the strategy should change, without modifying the downstream controller.

    \item We demonstrate body-grounded replanning under load and asymmetric mobility constraints in simulation and on a real robot, showing reduced physical execution cost and effective strategy adaptation, and further demonstrate the same interface on three contact-rich manipulation tasks.
\end{itemize}

\section{Related Work}
\label{sec:related_work}

Our work lies at the intersection of high-level manipulation planning, failure-aware replanning, and control from internal physical state.
We focus on a question between these areas: whether a high-level manipulation strategy should change in response to the robot's evolving physical condition.

\subsection{Externally Grounded Manipulation Planning}

Language-conditioned manipulation systems use visual observations, object geometry, and task semantics to select actions and strategies.
Prior work includes visuomotor policies~\cite{shridhar2021cliport}, language-model-guided skill selection~\cite{ichter2022saycan}, code-generated policies~\cite{codeaspolicies2022}, spatial affordance and value maps~\cite{huang2023voxposer}, and closed-loop language reasoning~\cite{huang2022inner}.
More recently, vision-language-action (VLA) models have learned mappings from visual observations and language instructions directly to robot actions using large-scale robot demonstrations~\cite{kim24openvla, intelligence2026pi07steerablegeneralistrobotic, nvidia2025gr00tn1openfoundation}.

These approaches primarily ground manipulation in the task and external scene.
Although learned policies may include proprioceptive state, it typically conditions action generation rather than explicitly evaluating whether the current high-level strategy remains physically appropriate.
Our approach adds this strategy-level body feedback: the task and scene can remain unchanged while the robot selects a different strategy according to its current physical condition.

\subsection{Failure Monitoring and Replanning}

Closed-loop manipulation systems monitor execution and revise plans based on feedback during execution.
Prior work has incorporated execution feedback into LLM-based planning~\cite{Bhat_2025}, used vision-language models for error detection and replanning~\cite{mei2024replanvlm}, and combined temporal consistency with VLM-based progress monitoring~\cite{AgiaSinhaEtAl2024}.
REFLECT~\cite{liu2023reflect} further summarizes multisensory robot experience to support LLM-based failure explanation and corrective planning.

These approaches focus primarily on identifying and correcting execution problems.
Our focus is instead whether an otherwise valid strategy remains physically compatible with the robot executing it.
A strategy may still make correct task progress while producing excessive joint effort or approaching a joint limit.
Body-grounded replanning can therefore reconsider an otherwise valid strategy
based on emerging physical difficulty and use the underlying physical state
to determine how the strategy should change.

\subsection{Internal Physical State in Robot Control and Planning}

Proprioceptive and force-related signals are fundamental to robot control, including stabilization, trajectory tracking, and interaction control~\cite{khatib1987operational,siciliano2010robotics}.
Physical constraints have also been incorporated into manipulation planning.
Constrained motion planners account for factors such as joint and torque limits when searching for feasible robot motions~\cite{berenson2009manipulation}, while forceful manipulation planning reasons over discrete strategies and continuous parameters subject to torque and friction constraints~\cite{holladay2021forceful,holladay2024robust}.

These methods show that physical constraints can affect both motion feasibility and manipulation strategy.
They reason over modeled constraints to search for physically feasible motions or strategies.
Our focus is complementary: we use internal physical state during execution to reassess whether the current high-level strategy remains physically appropriate.
This enables the robot to adapt its strategy as physical difficulty emerges, while preserving the task objective and downstream controller.

\section{Method}
\label{sec:method}

\begin{figure*}[t]
    \centering
    \includegraphics[width=\linewidth]{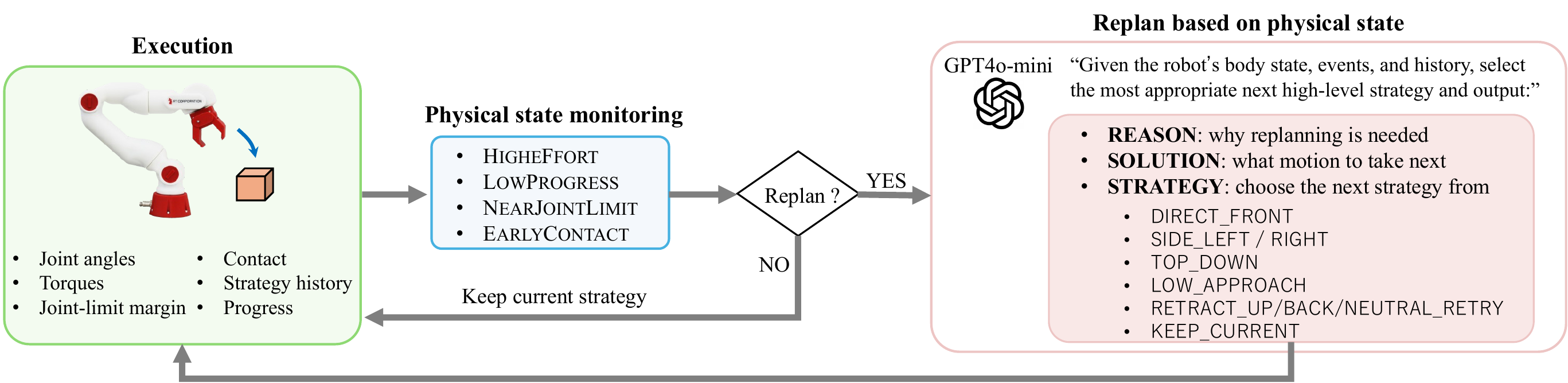}
    \caption{\namiko{changed font}
    Body-grounded high-level replanning: internal physical state triggers strategy reconsideration, and an LLM selects a context-dependent alternative.
    }
    \vspace{-3mm}
    \label{fig:pipeline}
\end{figure*}

\subsection{Problem Formulation and Overview}

We consider a manipulation task in which a robot pursues a fixed task goal using one of multiple high-level strategies.
Execution is indexed by discrete control step $t=0,1,\ldots$.
An environment-centered planner selects an initial strategy $g_0 \in \mathcal{G}$ based on the task goal and external environment, where $\mathcal{G}$ denotes the set of candidate strategies.
A fixed low-level controller then executes the selected strategy.

Our objective is to determine whether the current strategy $g_t$ remains physically appropriate as the robot's condition evolves and, when necessary, replace it with another strategy in $\mathcal{G}$ without changing the task goal or low-level controller.
We represent the robot's internal physical state as
\begin{equation}
    b_t = \{q_t,\dot{q}_t,\tau_t,c_t,p_t\},
\end{equation}
where $q_t$ and $\dot{q}_t$ are joint position and velocity, $\tau_t$ is joint effort, $c_t$ represents binary contact information, and $p_t$ denotes \revise{incremental} task progress.

Our framework separates \emph{when} a strategy should be reconsidered from \emph{how} it should change.
Measured body-state events provide the primary reactive evidence for strategy reconsideration, while a short-horizon predictor can additionally provide anticipatory evidence of impending difficulty.
When either triggers reconsideration, an LLM interprets the underlying joint-level state together with recent body-state statistics, and strategy history to select a context-dependent alternative from $\mathcal{G}$.
The task goal and low-level controller remain unchanged throughout replanning.
Figure~\ref{fig:pipeline} illustrates the framework, and Algorithm~\ref{alg:pipeline} summarizes the execution loop.

\begin{algorithm}[t]
\caption{Body-grounded high-level replanning}
\label{alg:pipeline}
\begin{algorithmic}[1]
\Require task goal, strategy set $\mathcal{G}$,
event detector $\mathcal{D}$, predictor $F$, replan budget $K$
\State select initial strategy $g \in \mathcal{G}$ from the task goal
and external environment
\State $\mathcal{H} \leftarrow [\,]$;\; $k \gets 0$
\While{task not finished}
    \State execute strategy $g$ and observe body state $b_t$
    \State $s_t \gets \Call{BodyStatistics}{b_{t-W+1:t}}$
    \State $(\hat q_{t+\Delta},\hat{\tau}_{t+\Delta})
    \gets F(q_{t-H+1:t},\dot q_{t-H+1:t},\tau_{t-H+1:t})$
    \State $E_t \gets
    \mathcal{D}(b_{t-W+1:t},s_t,
    \hat q_{t+\Delta},\hat{\tau}_{t+\Delta})$
    \If{$E_t \neq \{\textsc{Normal}\}$ \textbf{and} $k < K$}
        \State $g_{\mathrm{new}} \gets
        \pi_{\mathrm{LLM}}
        (g,b_t,s_t,E_t,\mathcal{H},\mathcal{G})$
        \State append $g$ to $\mathcal{H}$
        \State $g \gets g_{\mathrm{new}}$;\; $k \gets k+1$
    \EndIf
\EndWhile
\end{algorithmic}
\end{algorithm}

\subsection{When to Replan: Body-State Monitoring and Prediction}
\label{sec:event_detection}
\namiko{changed subsection title to correspond to "When"}
The event detector determines when the current high-level strategy should be reconsidered.
To suppress transient fluctuations, we summarize recent execution using rolling statistics $s_t$ over the latest $W$ control steps.
These include mean joint-effort norm, mean task progress, minimum joint-limit margin, contact count, and per-joint effort statistics.
Task progress at each control step is defined as the reduction in distance to the task target,
\begin{equation}
    p_t = d_{t-1} - d_t,
\end{equation}
where $d_t$ is the distance to the target at step $t$.
Mean task progress is the average of $p_t$ over the rolling window.

We consider four body-state events:
\textsc{HighEffort}, \textsc{LowProgress}, \textsc{NearJointLimit}, and \textsc{EarlyContact}.
They respectively indicate sustained physical effort, insufficient task progress, proximity to a joint limit, and contact before the intended interaction phase.
Measured events must persist for a predefined number of control steps before triggering replanning, and a cooldown period prevents rapid strategy switching.

In addition to these measured events, a short-horizon predictor provides anticipatory evidence for \textsc{HighEffort} and \textsc{NearJointLimit}.
Given the previous $H$ control steps of joint positions, velocities, and efforts, the predictor $F$ estimates joint position and effort $\Delta$ control steps ahead:
\begin{equation}
\begin{aligned}
(\hat q_{t+\Delta},\hat{\tau}_{t+\Delta})
= F(q_{t-H+1:t},\ 
    \dot q_{t-H+1:t},\ \tau_{t-H+1:t}),
\end{aligned}
\end{equation}
where $H$ is the predictor input-history length and $\Delta$ is the prediction
horizon.
The same event criteria are applied to the measured and predicted states, and either can trigger strategy reconsideration.

Importantly, prediction affects only the evidence used to trigger reconsideration; it does not determine which strategy is selected.
Once triggered, strategy selection uses the body state and execution context described in the following section.
The predictor model and training  data are described in Sec.~\ref{sec:implementation}.

\subsection{High-Level Strategy Space}
\label{sec:strategy_space}

The strategy set $\mathcal{G}$ contains \revise{a fixed set of} coarse manipulation behaviors that change how the robot approaches or retreats from the task while preserving the task goal.
It includes five approach strategies:
\texttt{DIRECT\_FRONT},
\texttt{SIDE\_LEFT},
\texttt{SIDE\_RIGHT},
\texttt{TOP\_DOWN}, and
\texttt{LOW\_APPROACH};
three retreat strategies:
\texttt{RETRACT\_BACK},
\texttt{RETRACT\_UP}, and
\texttt{RETRACT\_NEUTRAL\_RETRY};
and \texttt{KEEP\_CURRENT}.

Approach strategies define target-relative intermediate waypoints, whereas retreat strategies create clearance before a subsequent approach.
\texttt{KEEP\_CURRENT} allows the current strategy to continue when switching is unnecessary.
\revise{The LLM is restricted to selecting a strategy from $\mathcal{G}$ rather than generating arbitrary behaviors, ensuring compatibility with the downstream controller and providing a well-defined interface between high-level reasoning and low-level execution.}
The rule-based baseline and our method share the same strategy set and low-level controller; they differ only in how the next strategy is selected.
\subsection{How to Replan: Body-Grounded Strategy Selection}
\label{sec:body_reasoning}
\namiko{changed subsection title to correspond to "How"}
When a measured or predicted body-state event triggers reconsideration, the LLM selects the next high-level strategy from $\mathcal{G}$:
\begin{equation}
g_{t+1} =
\pi_{\mathrm{LLM}}
\bigl(
g_t,b_t,s_t,E_t,\mathcal{H}_t,\mathcal{G}
\bigr),
\end{equation}
where $E_t$ denotes the detected body-state events and $\mathcal{H}_t$ contains the previously attempted strategies.
The event labels summarize why reconsideration was triggered, while $b_t$ and $s_t$ provide the underlying current and recent physical evidence.

The LLM receives the current strategy, per-joint positions, efforts, and joint-limit margins, recent body-state statistics, detected events, previously attempted strategies, and the available strategy set.
An event alone does not uniquely determine the appropriate response.
For example, \textsc{NearJointLimit} indicates that the current motion is approaching a kinematic constraint, but the appropriate alternative depends on which joint is constrained, the current configuration, and strategies already attempted.
The LLM therefore uses the underlying physical state and execution context to select a context-dependent alternative rather than applying a fixed event-to-strategy mapping.

We implement $\pi_{\mathrm{LLM}}$ using GPT-4o-mini~\cite{openai2024gpt4omini}\namiko{added citation}.
The prompt provides qualitative guidance on the physical meaning of each event and the characteristics of each strategy, but does not prescribe a fixed event-to-strategy mapping.
The output is structured into three fields: \texttt{REASON}, \texttt{SOLUTION}, and \texttt{STRATEGY}.
\texttt{REASON} identifies the physical issue from the provided body-state evidence, \texttt{SOLUTION} describes the corresponding high-level response, and \texttt{STRATEGY} maps that response to one of the available discrete strategies.
This structure makes the intermediate interpretation and intended response explicit before the final strategy selection, which also allows us to inspect whether the selected strategy is consistent with the supplied physical evidence.
Only \texttt{STRATEGY} is passed to the robot for execution.
Figure~\ref{fig:prompt} summarizes the prompt structure.

\begin{figure}[t]
\centering
\fbox{\begin{minipage}{0.94\linewidth}
\footnotesize
You are a replanner for robotic manipulation.
The robot is moving its arm toward a target object.
Given information about the robot's physical state, including joint angles,
per-joint torques, joint-limit margins, and task progress, select the single
best strategy to execute next.

\textbf{Strategy descriptions:}
\begin{itemize}
    \item \texttt{DIRECT\_FRONT}: Approach the target directly from the front.
    This is the shortest approach, but may encounter joint limits depending on
    the robot configuration.

    \item \texttt{SIDE\_LEFT}: Approach from the left side.
    Useful when the robot configuration is biased toward the right.

    \item \texttt{SIDE\_RIGHT}: Approach from the right side.
    Useful when the robot configuration is biased toward the left.

    \item \texttt{TOP\_DOWN}: Approach from above.
    This tends to reduce joint load but requires a longer path.

    \item \texttt{LOW\_APPROACH}: Approach through a waypoint in front of and
    below the target.
    This provides an alternative direction when an upper approach is difficult.

    \item \texttt{RETRACT\_BACK}: Retract the arm toward the robot, away from
    the target.
    This strategy performs only the retreat; re-approach requires a subsequent
    strategy.

    \item \texttt{RETRACT\_UP}: Retract to a position in front of and above
    the target.
    This strategy performs only the retreat and is useful for creating
    clearance after contact.

    \item \texttt{RETRACT\_NEUTRAL\_RETRY}: Return to a neutral configuration
    and retry.
    Use this as a fallback when the robot is stuck.

    \item \texttt{KEEP\_CURRENT}: Maintain the current strategy to avoid
    unnecessary strategy switching.
\end{itemize}

\textbf{Decision guidelines:}~\namiko{changed font bellow}
\begin{itemize}
    \item \textsc{HighEffort}: High joint torque indicates that the current
    direction is physically demanding; consider an alternative direction.

    \item \textsc{LowProgress}: The current strategy is not making sufficient
    progress toward the target.

    \item \textsc{NearJointLimit}: Continuing the current motion may limit
    further movement; consider retreating or changing direction.

    \item \textsc{EarlyContact}: Unexpected contact has occurred before the
    intended interaction; approaching from above or from the opposite side may
    be effective.
\end{itemize}

\textbf{Important:}
Avoid changing strategies merely for the sake of trying a different one.
Switch only when the available physical evidence suggests that another
strategy is likely to improve execution; otherwise select
\texttt{KEEP\_CURRENT}.

\textbf{Output format:}
Output exactly the following three lines:

\texttt{REASON: <1--2 sentences explaining the reason for replanning>}\\
\texttt{SOLUTION: <1--2 sentences summarizing the next action>}\\
\texttt{STRATEGY: <one strategy name>}

The \texttt{STRATEGY} field must contain exactly one of:
\texttt{DIRECT\_FRONT},
\texttt{SIDE\_LEFT},
\texttt{SIDE\_RIGHT},
\texttt{TOP\_DOWN},
\texttt{LOW\_APPROACH},
\texttt{RETRACT\_BACK},
\texttt{RETRACT\_UP},
\texttt{RETRACT\_NEUTRAL\_RETRY}, or
\texttt{KEEP\_CURRENT}.
\end{minipage}}
\caption{The prompt for body-grounded strategy selection. Numerical values are inserted at runtime. Angle-bracketed text denotes output fields to be completed by the LLM.}
\label{fig:prompt}
\end{figure}

\section{Experimental Setup}
\label{sec:experimental_setup}

We evaluate body-grounded replanning under controlled variations in the robot's physical condition using reaching tasks in simulation and on hardware, followed by demonstrations on three additional manipulation tasks.

\subsection{Robot Platforms}
\label{sec:platforms}

Simulation experiments use a Franka Panda 7-DoF manipulator in robosuite~\cite{robosuite2020} with MuJoCo~\cite{todorov2012mujoco}.
Hardware experiments use a CRANE-X7 7-DoF manipulator with Dynamixel actuators controlled through MoveIt~\cite{coleman2014moveit}\namiko{added citations}.
Joint state and effort are recorded at $20$\,Hz on both platforms.

\subsection{Implementation Details}
\label{sec:implementation}
Body-state statistics are computed over a recent window of $W=15$ control steps ($0.75$\,s).
Measured events must persist for $10$ steps ($0.50$\,s) before triggering replanning, and a $10$-step cooldown is used to prevent repeated triggers.

\revise{For anticipatory detection, the predictor uses the previous $H=20$ control steps ($1.0$\,s) of joint positions, velocities, and efforts to predict joint positions and efforts $\Delta=10$ steps ($0.50$\,s) ahead.
We use a linear Ridge regression model trained on 480 random-target reaching trajectories collected under nominal conditions and three added-load configurations, with the load attached at different locations on the arm.}
The LLM replanner uses GPT-4o-mini with temperature $0$.\namiko{deleted line break in this paragraph}

\subsection{Compared Methods}
\label{sec:baselines}

We compare four methods that share the same low-level controller and candidate strategy set, differing only in when and how the strategy is changed.

\paragraph{Fixed Planning (Fixed)}
Fixed executes the initial strategy without replanning or body-state feedback.

\paragraph{Environment-only Body-Agnostic Replanning (Env.)}
Env.\ replans without using body-state information.
In simulation, Env. cycles through a fixed strategy sequence every $80$ control steps ($4.0$\,s). 
On hardware, it advances the sequence after three unsuccessful MoveIt motion executions; a safety effort stop can also trigger advancement.
This baseline isolates the effect of replanning without body grounding.

\paragraph{Rule-Based Body Replanning (Rule)}
Rule uses the same measured and predicted body-state events as Ours, but selects strategies using hand-designed event-specific priorities together with the current strategy and attempt history.
\textsc{HighEffort} and \textsc{EarlyContact} prioritize \textsc{RetractUp}, \textsc{RetractBack}, \textsc{LowApproach}, and then an untried approach direction; 
\textsc{NearJointLimit} prioritizes \textsc{RetractBack} and then \textsc{RetractNeutralRetry}; 
and \textsc{LowProgress} prioritizes an untried \textsc{RetractUp} or \textsc{LowApproach} before an untried side approach.
These priorities select the next untried candidate rather than defining a sequence of consecutive actions.
Rule does not use the underlying continuous joint-level state.

\paragraph{Body-Grounded Replanning (Ours)}
Ours uses the underlying joint-level state, recent execution statistics, and strategy history to select among the same candidate strategies using the LLM replanner.
Thus, the Rule--Ours comparison isolates fixed event-based responses from context-dependent strategy selection grounded in continuous body state.

\subsection{Controlled Reaching Evaluation}
\label{sec:controlled_tasks}

We evaluate the methods on the same reaching task while systematically varying the robot's physical condition.
The evaluation comprises ten conditions: three load and seven asymmetric mobility conditions.

\paragraph{Load.}
Additional mass is attached at the wrist, forearm, or upper arm. We use $500$\,g in simulation and $250$\,g on hardware.

\paragraph{Asymmetric Mobility.}
One joint is restricted at a time, yielding seven conditions.
In simulation, each tested joint is restricted to $70\%$ of its nominal range.
On hardware, Joints~1--5 are restricted to $60\%$ and Joints~6--7 to $80\%$ to preserve feasible execution.

For each simulation condition, we generate $100$ episodes with varied target positions and initial joint configurations, including joint-wise perturbations.
For each episode index, all four methods use the same target position, initial joint configuration, and initial strategy, enabling paired comparison.
Each episode has an execution budget of $250$ control steps, corresponding to a maximum recorded duration of $12.45$\,s.

On hardware, we conduct $8$ trials per condition for Env., Rule, and Ours with a $45$\,s execution budget.
Fixed is evaluated systematically in simulation but omitted from the hardware comparison because its available trials were not collected under the same $45$\,s protocol.

\subsection{Evaluation Metrics}
\label{sec:metrics}

We report task success, physical effort, and active execution time.
Success denotes task completion within the execution budget.

\paragraph{Physical Effort.}
In simulation, we report the mean joint-torque norm over active execution,
\begin{equation}
C_{\tau}
=
\frac{1}{N}
\sum_{t=1}^{N}\lVert\tau_t\rVert_2 ,
\end{equation}
where $N$ is the number of active control steps.
On hardware, we analogously report the mean norm of raw Dynamixel actuator-effort readings.
Because these readings are not calibrated joint torques, hardware effort is compared only within the real platform.

\paragraph{Execution and Completion Time.}
Active execution time excludes LLM inference pauses on both platforms.
In simulation, it is computed from the number of control steps at $20$\,Hz; on hardware, it is measured during active robot execution.

For both platforms, \revise{to compare completion efficiency without conditioning on successful trials, we use restricted completion time over all trials.}~\namiko{add explanation for restricted completion time:}
Successful trials use their active execution time, while failures are assigned the corresponding execution horizon: $12.45$\,s in simulation and $45$\,s on hardware.
Lower values therefore reflect both faster and more reliable completion.

\paragraph{Statistical Analysis.}
Simulation comparisons between Ours and Rule or Env.\ use paired two-sided Wilcoxon signed-rank tests over the $100$ matched episodes for joint torque and restricted completion time.
Holm correction is applied across the ten conditions separately for each baseline--metric comparison, and rank-biserial correlation is reported as the effect size.

\subsection{Additional Manipulation Tasks}
\label{sec:additional_tasks}

We additionally demonstrate the framework on three contact-rich real-robot tasks: obstacle-avoiding button pressing, box-lid opening, and lint-roller extraction (Fig.~\ref{fig:real_robot_tasks}).
All tasks retain the same body-event semantics, strategy vocabulary, and structured LLM output, while task-specific motion and event parameters are configured separately.
These experiments demonstrate reuse of the body-grounded replanning interface across different manipulation tasks.

\section{Results and Discussion}
\label{sec:results}

\subsection{Simulation Results}
\begin{table*}[t]
\centering
\caption{
Simulation results over $100$ paired episodes per condition.
Torque is mean$\pm$std over episodes; restricted completion time assigns $12.45$\,s to failures and excludes LLM-inference pauses.
Mean rows average the corresponding condition-wise means.
\textbf{Bold} indicates the best value per row and metric.
}
\label{tab:sim_results}
\footnotesize
\setlength{\tabcolsep}{2pt}
\begin{tabular}{lcccccccccccc}
\toprule
&
\multicolumn{4}{c}{Success (\%) $\uparrow$} &
\multicolumn{4}{c}{Mean joint-torque norm $\downarrow$} &
\multicolumn{4}{c}{Restricted time [s] $\downarrow$} \\
\cmidrule(lr){2-5}
\cmidrule(lr){6-9}
\cmidrule(lr){10-13}
Condition &
Fixed & Env. & Rule & Ours &
Fixed & Env. & Rule & Ours &
Fixed & Env. & Rule & Ours \\
\midrule

\multicolumn{13}{l}{
\textbf{Load} \textit{($+500$\,g)}
} \\

Wrist
& 99 & \textbf{100} & 91 & 98
& $52.8\pms{5.7}$ & $52.6\pms{5.8}$ & $52.6\pms{9.2}$ & $\mathbf{49.5\pms{7.4}}$
& $\mathbf{3.21\pms{1.39}}$ & $3.35\pms{1.48}$ & $3.47\pms{3.09}$ & $3.38\pms{2.27}$ \\

Forearm
& 99 & \textbf{100} & 91 & 99
& $51.4\pms{5.7}$ & $51.4\pms{5.6}$ & $51.2\pms{9.3}$ & $\mathbf{48.4\pms{7.7}}$
& $3.24\pms{1.49}$ & $3.33\pms{1.46}$ & $3.44\pms{3.09}$ & $\mathbf{3.13\pms{2.00}}$ \\

Upper arm
& 99 & \textbf{100} & 92 & 97
& $49.7\pms{5.6}$ & $49.9\pms{5.4}$ & $49.8\pms{9.3}$ & $\mathbf{46.9\pms{7.3}}$
& $\mathbf{3.24\pms{1.48}}$ & $3.28\pms{1.42}$ & $3.32\pms{2.96}$ & $3.45\pms{2.58}$ \\

\cmidrule(lr){1-13}

\textit{Mean}
& 99.0 & \textbf{100.0} & 91.3 & 98.0
& $51.3$ & $51.3$ & $51.2$ & $\mathbf{48.3}$
& $\mathbf{3.23}$ & $3.32$ & $3.41$ & $3.32$ \\

\midrule

\multicolumn{13}{l}{
\textbf{Asymmetric mobility} \textit{(single-joint range factor $0.70$)}
} \\

Joint 1
& 99 & \textbf{100} & 91 & 97
& $48.8\pms{5.4}$ & $48.8\pms{5.4}$ & $49.3\pms{9.2}$ & $\mathbf{45.8\pms{7.0}}$
& $3.21\pms{1.40}$ & $3.33\pms{1.48}$ & $3.43\pms{3.07}$ & $\mathbf{3.17\pms{2.32}}$ \\

Joint 2
& 78 & \textbf{93} & 74 & 92
& $50.8\pms{4.9}$ & $49.7\pms{5.1}$ & $49.3\pms{9.3}$ & $\mathbf{46.1\pms{7.1}}$
& $5.47\pms{3.83}$ & $4.72\pms{2.64}$ & $5.15\pms{4.46}$ & $\mathbf{4.02\pms{3.01}}$ \\

Joint 3
& 99 & \textbf{100} & 91 & 99
& $48.8\pms{5.4}$ & $48.8\pms{5.3}$ & $49.0\pms{9.1}$ & $\mathbf{46.1\pms{7.2}}$
& $3.21\pms{1.40}$ & $3.34\pms{1.52}$ & $3.46\pms{3.08}$ & $\mathbf{3.01\pms{1.89}}$ \\

Joint 4
& 99 & \textbf{100} & 82 & 90
& $49.2\pms{5.7}$ & $49.5\pms{5.9}$ & $47.2\pms{10.0}$ & $\mathbf{45.1\pms{6.8}}$
& $\mathbf{2.83\pms{1.60}}$ & $2.91\pms{1.72}$ & $4.35\pms{4.02}$ & $4.03\pms{3.32}$ \\

Joint 5
& 99 & \textbf{100} & 93 & 98
& $48.9\pms{5.2}$ & $49.0\pms{5.0}$ & $49.0\pms{8.9}$ & $\mathbf{45.8\pms{7.0}}$
& $\mathbf{3.21\pms{1.41}}$ & $3.25\pms{1.32}$ & $3.30\pms{2.84}$ & $3.25\pms{2.18}$ \\

Joint 6
& 99 & \textbf{100} & 88 & 99
& $48.6\pms{5.3}$ & $48.8\pms{5.5}$ & $50.6\pms{10.1}$ & $\mathbf{46.4\pms{6.7}}$
& $\mathbf{3.05\pms{1.20}}$ & $3.16\pms{1.40}$ & $4.01\pms{3.61}$ & $3.19\pms{1.93}$ \\

Joint 7
& 99 & \textbf{100} & 92 & 98
& $48.7\pms{5.5}$ & $48.8\pms{5.3}$ & $49.0\pms{9.0}$ & $\mathbf{46.1\pms{7.1}}$
& $3.30\pms{1.56}$ & $3.33\pms{1.54}$ & $3.39\pms{2.94}$ & $\mathbf{3.15\pms{2.26}}$ \\

\cmidrule(lr){1-13}

\textit{Mean}
& 96.0 & \textbf{99.0} & 87.3 & 96.1
& $49.1$ & $49.0$ & $49.1$ & $\mathbf{45.9}$
& $3.47$ & $3.43$ & $3.87$ & $\mathbf{3.40}$ \\

\midrule
\textit{Overall Mean}
& 96.9 & \textbf{99.3} & 88.5 & 96.7
& 49.8 & 49.7 & 49.7 & \textbf{46.6}
& 3.40 & 3.40 & 3.73 & \textbf{3.38} \\

\bottomrule
\end{tabular}
\end{table*}
Ours maintains high task success ($90$--$99\%$) while achieving the lowest mean joint-torque norm in all ten conditions (Table~\ref{tab:sim_results}).
Paired Wilcoxon tests with Holm correction confirm significantly lower torque than both Rule and Env.\ in every condition, with moderate-to-large effect sizes relative to Rule ($|r|=0.38$--$0.64$) and large effects relative to Env.\ ($|r|=0.66$--$0.87$).
The reduction is consistent across both load and asymmetric-mobility conditions, showing that body-grounded replanning reduces physical execution cost while preserving high task success.

The comparison with Rule isolates the importance of how body-state information
is used for strategy selection.
Both methods receive the same measured and predicted events and select from the same candidate strategies, but Rule applies hand-designed event-based responses whereas Ours reasons over the underlying joint-level state.
Ours achieves higher overall success ($96.7\%$ vs.\ $88.5\%$) and lower restricted completion time in all seven asymmetric mobility conditions.
This suggests that detecting physical difficulty alone is not sufficient; the underlying physical state provides useful context for selecting an appropriate alternative strategy.

\subsection{Real-Robot Results}

On hardware, Ours and Env.\ achieve the same overall success rate of $96.3\%$, substantially higher than Rule at $66.3\%$ (Table~\ref{tab:real_results}).
Despite matching Env.\ in overall success, Ours achieves lower restricted completion time in nine of ten conditions and the lowest average across all conditions ($24.57$\,s vs.\ $28.18$\,s for Env.).
This indicates that body-agnostic replanning can achieve high success by systematically trying alternatives, whereas body-grounded replanning generally reaches successful strategies more efficiently.

The comparison with Rule further highlights the importance of how body-state information is used for strategy selection.
Rule and Ours receive the same measured and predicted body-state events and select from the same candidate strategies, but Rule relies on hand-designed event-based responses.
Ours improves overall success from $66.3\%$ to $96.3\%$ and achieves lower restricted completion time in all ten conditions ($24.57$\,s vs.\ $35.40$\,s on average).
Together with the simulation results, this supports using the underlying joint-level state to determine how the strategy should change, rather than mapping detected physical events directly to fixed responses.

Hardware effort shows a less uniform condition-wise trend than simulated joint torque, with Ours achieving the lowest mean actuator-effort norm in five of ten conditions.
Nevertheless, Ours has the lowest average effort across the ten hardware conditions ($2.18$ vs.\ $2.28$ for Env.\ and $2.33$ for Rule).
Overall, the hardware results show that body-grounded replanning maintains high task success while enabling efficient, context-dependent strategy adaptation across changing physical conditions.

\begin{table*}[t]
\centering
\caption{
Real-robot results over $8$ trials per method and condition.
Actuator effort is mean$\pm$std over trials; restricted completion time assigns $45$\,s to failures.
Mean rows average the corresponding condition-wise means.
\textbf{Bold} indicates the best value per row and metric.
}
\label{tab:real_results}
\footnotesize
\setlength{\tabcolsep}{2.8pt}
\begin{tabular}{lccccccccc}
\toprule
&
\multicolumn{3}{c}{Success (\%) $\uparrow$} &
\multicolumn{3}{c}{Mean actuator-effort norm $\downarrow$} &
\multicolumn{3}{c}{Restricted time [s] $\downarrow$} \\
\cmidrule(lr){2-4}
\cmidrule(lr){5-7}
\cmidrule(lr){8-10}
Condition &
Env. & Rule & Ours &
Env. & Rule & Ours &
Env. & Rule & Ours \\
\midrule

\multicolumn{10}{l}{
\textbf{Load} \textit{($+250$\,g)}
} \\

Wrist
& \textbf{100} & 75 & \textbf{100}
& $2.53\pms{0.15}$ & $2.33\pms{0.16}$ & $\mathbf{2.26\pms{0.26}}$
& $32.74\pms{5.56}$ & $33.03\pms{10.99}$ & $\mathbf{28.79\pms{4.80}}$ \\

Forearm
& 75 & 75 & \textbf{100}
& $2.43\pms{0.10}$ & $\mathbf{2.02\pms{0.37}}$ & $2.03\pms{0.33}$
& $34.94\pms{8.79}$ & $28.49\pms{12.62}$ & $\mathbf{27.07\pms{8.26}}$ \\

Upper arm
& \textbf{87.5} & 62.5 & \textbf{87.5}
& $2.38\pms{0.19}$ & $1.97\pms{0.11}$ & $\mathbf{1.86\pms{0.35}}$
& $31.35\pms{7.70}$ & $32.84\pms{13.64}$ & $\mathbf{28.51\pms{10.18}}$ \\

\cmidrule(lr){1-10}

\textit{Mean}
& 87.5 & 70.8 & \textbf{95.8}
& $2.45$ & $2.11$ & $\mathbf{2.05}$
& $33.01$ & $31.45$ & $\mathbf{28.12}$ \\

\midrule

\multicolumn{10}{l}{
\textbf{Asymmetric mobility}
} \\

Joint 1
& \textbf{100} & 62.5 & \textbf{100}
& $\mathbf{2.22\pms{0.29}}$ & $2.44\pms{0.22}$ & $2.34\pms{0.29}$
& $28.85\pms{4.48}$ & $32.94\pms{11.00}$ & $\mathbf{18.76\pms{5.16}}$ \\

Joint 2
& \textbf{100} & 75 & \textbf{100}
& $2.29\pms{0.18}$ & $2.33\pms{0.19}$ & $\mathbf{2.12\pms{0.25}}$
& $27.81\pms{5.96}$ & $35.60\pms{9.08}$ & $\mathbf{23.77\pms{6.83}}$ \\

Joint 3
& \textbf{100} & \textbf{100} & \textbf{100}
& $\mathbf{2.30\pms{0.25}}$ & $2.37\pms{0.25}$ & $2.55\pms{0.16}$
& $27.01\pms{8.10}$ & $33.82\pms{7.89}$ & $\mathbf{25.32\pms{3.61}}$ \\

Joint 4
& \textbf{100} & 62.5 & \textbf{100}
& $\mathbf{2.17\pms{0.18}}$ & $2.34\pms{0.27}$ & $2.23\pms{0.45}$
& $26.21\pms{3.68}$ & $36.56\pms{9.22}$ & $\mathbf{19.52\pms{4.66}}$ \\

Joint 5
& \textbf{100} & 50 & \textbf{100}
& $2.20\pms{0.20}$ & $2.48\pms{0.20}$ & $\mathbf{2.09\pms{0.27}}$
& $19.24\pms{3.77}$ & $38.74\pms{9.54}$ & $\mathbf{17.40\pms{7.51}}$ \\

Joint 6
& \textbf{100} & 37.5 & \textbf{100}
& $\mathbf{2.02\pms{0.38}}$ & $2.52\pms{0.34}$ & $2.21\pms{0.30}$
& $\mathbf{22.12\pms{5.72}}$ & $40.26\pms{7.36}$ & $26.24\pms{6.44}$ \\

Joint 7
& \textbf{100} & 62.5 & 75
& $2.30\pms{0.34}$ & $2.48\pms{0.25}$ & $\mathbf{2.13\pms{0.25}}$
& $31.57\pms{7.75}$ & $41.69\pms{3.19}$ & $\mathbf{30.34\pms{10.02}}$ \\

\cmidrule(lr){1-10}

\textit{Mean}
& \textbf{100.0} & 64.3 & 96.4
& $\mathbf{2.21}$ & $2.42$ & $2.24$
& $26.69$ & $37.09$ & $\mathbf{23.05}$ \\

\midrule
\textit{Overall Mean}
& \textbf{96.3} & 66.3 & \textbf{96.3}
& 2.28 & 2.33 & \textbf{2.18}
& 28.18 & 35.40 & \textbf{24.57} \\

\bottomrule
\end{tabular}
\end{table*}
\vspace{-2mm}

\subsection{Additional Manipulation Tasks}
\begin{figure*}[t]
\centering
\includegraphics[width=\textwidth]{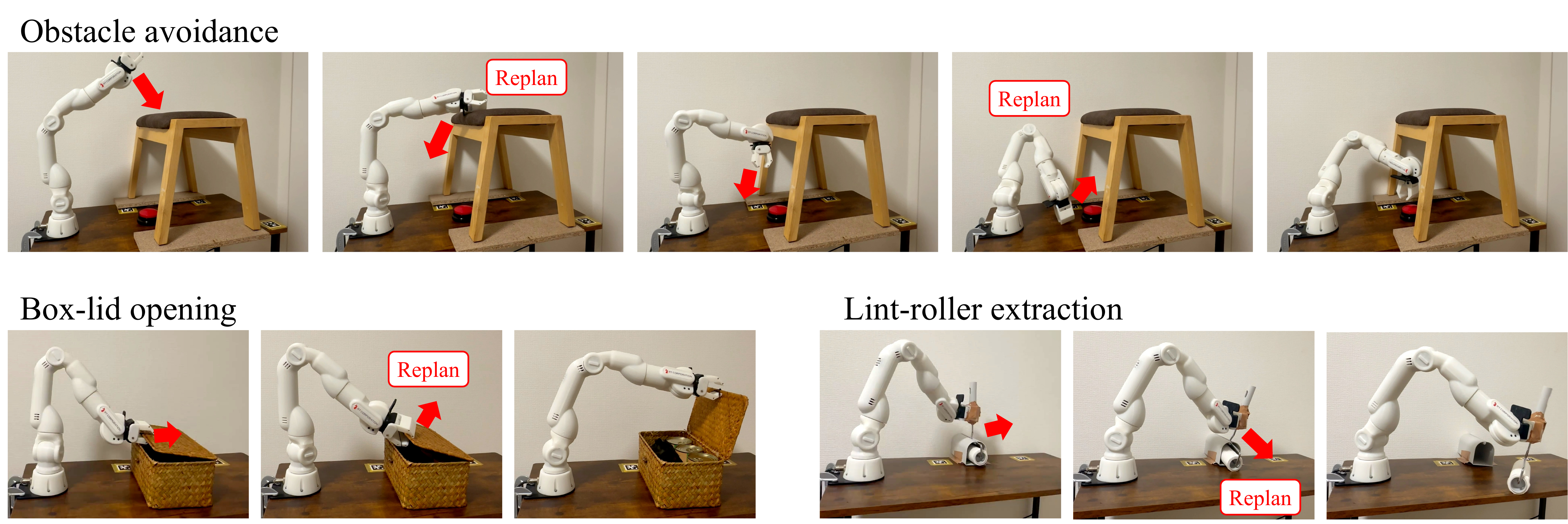}
\caption{Additional real-robot demonstrations: obstacle-avoiding button pressing, box-lid opening, and lint-roller extraction. \revise{The robot changes its strategy in response to the encountered physical difficulty and completes the task.}}
\vspace{-2mm}
\label{fig:real_robot_tasks}
\end{figure*}

We further apply the same body-grounded replanning interface to three
contact-rich tasks: obstacle-avoiding button pressing, box-lid opening,
and lint-roller extraction.~\namiko{added 2 paragraphs bellow}
Figure~\ref{fig:real_robot_tasks} shows representative executions in which unexpected contact or impeded motion triggers replanning.
The robot changes its strategy in response to the encountered physical difficulty and subsequently completes the task.

\begin{figure*}[t]
    \centering
    \includegraphics[width=0.94\textwidth]{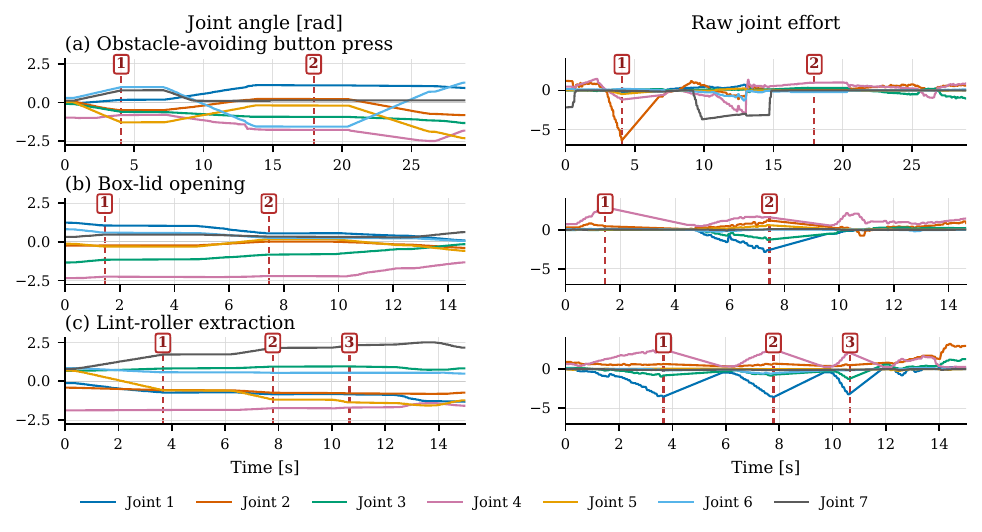}
    \caption{
    Body-state signals and replanning decisions during three additional real-robot manipulation tasks.
    \revise{Numbered markers indicate replanning decisions within each task.}
    \revise{Replanning changes the strategy in response to physical difficulty, which can reduce joint effort or restore task progress.}
    }
    \label{fig:real_task_signals}
\end{figure*}

Figure~\ref{fig:real_task_signals} shows the corresponding evolution of body-state signals around replanning decisions.
Across the tasks, strategy changes reduce elevated joint effort or restore task progress when the current strategy becomes physically unsuitable.
For example, replanning alleviates elevated effort during obstacle avoidance and box-lid opening, while repeated adaptation during lint-roller extraction allows execution to continue as the physical interaction changes.
Together, these demonstrations show that the same body-grounded interface can adapt strategies across different forms of physical interaction.

\subsection{Discussion}

Overall, the results support the distinction between geometric feasibility and physical suitability.
Env.\ shows that replanning alone can achieve high success by trying alternative strategies, but does not use the robot's physical condition to guide the choice.
Rule incorporates body-state events, yet its hand-designed responses do not account for the underlying continuous joint configuration.
Ours combines these two stages: body-state events determine when the current strategy should be reconsidered, while the underlying physical state informs how it should change.
The consistent advantage over Rule therefore suggests that detecting physical difficulty alone is insufficient; effective adaptation also requires reasoning about the physical context in which that difficulty occurs.

The benefit of body grounding extends beyond task success.
In simulation, Ours consistently reduces joint-torque norm even when other methods already reach the same task-space goal, showing that geometrically feasible strategies can differ substantially in their physical suitability.
On hardware, Ours maintains high success while achieving the lowest overall restricted completion time, demonstrating that body-grounded selection can also make adaptation more efficient.
Together with the additional manipulation tasks, these results show that internal physical state can inform high-level decisions about how a manipulation task should be performed.

\section{Conclusion}
\label{sec:conclusion}

We introduced \emph{body-grounded high-level replanning}, a framework that uses the robot's internal physical state to adapt manipulation strategies during execution.
Rather than using proprioceptive information only for low-level control, our approach uses body-state events to determine when a strategy should be reconsidered and the underlying joint-level state and execution history to determine how it should change, while leaving the task objective and low-level controller unchanged.

Controlled reaching experiments demonstrate that body grounding provides benefits beyond task success alone.
In simulation, our method achieves the lowest mean joint-torque norm across all ten load and asymmetric-mobility conditions while maintaining high task success.
On hardware, it achieves $96.3\%$ overall success and the lowest overall restricted completion time, while substantially outperforming rule-based body replanning in success.
The comparison with Rule further shows that detecting physical difficulty alone is not sufficient: using the underlying body state enables more effective selection of an alternative strategy.
Additional experiments on button pressing, box-lid opening, and lint-roller extraction demonstrate the applicability of the same body-grounded replanning interface across different forms of physical interaction.

More broadly, our results show that geometric feasibility does not imply physical suitability: effective manipulation requires reasoning not only about what is feasible in the world, but also about how it should be performed by the body executing it.
The current study uses a \revise{manually defined} discrete strategy space and relatively short-horizon tasks.
Future work will extend the framework to learned strategy spaces and more complex, long-horizon manipulation.

\section*{Acknowledgments}
This research was partially supported by JSPS KAKENHI Grant Numbers 24K17248 and 26K02996.

\bibliographystyle{unsrtnat}
\bibliography{references}

\end{document}